\documentclass[journal,trans]{IEEEtran}

\usepackage{epsfig}
\usepackage{algorithm} 
\usepackage{algorithmic} 
\usepackage{amsmath}
\usepackage{xcolor}
\usepackage{amsmath,epsfig,graphicx}
\usepackage{amsfonts}
\usepackage{bigstrut,multirow}
\usepackage{threeparttable}

\usepackage[colorlinks,linkcolor=red]{hyperref}
\usepackage{color}
\usepackage[switch]{lineno}

\begin{document}

\title{Iterative Manifold Embedding Layer
Learned by Incomplete Data
for Large-scale Image Retrieval}

\author{\IEEEauthorblockN{Jian Xu,
Chunheng Wang,
Chengzuo Qi,
Cunzhao Shi,
and
Baihua Xiao}


}

\markboth{Journal of \LaTeX\ Class Files,~Vol.~14, No.~8, August~2015}%
{Shell \MakeLowercase{\textit{et al.}}: Bare Demo of IEEEtran.cls for IEEE Transactions}

\IEEEtitleabstractindextext{%
\begin{abstract}

Existing manifold learning methods are not appropriate for image retrieval task, because most of them are unable to process query image and they have much additional computational cost especially for large scale database.

Therefore, we propose the iterative manifold embedding (IME) layer, of which the weights are learned off-line by unsupervised strategy, to explore the intrinsic manifolds by incomplete data.
On the large scale database that contains 27000 images, IME layer is more than 120 times faster than other manifold learning methods to embed the original representations at query time.

We embed the original descriptors of database images which lie on manifold in a high dimensional space into manifold-based representations iteratively to generate the IME representations  in off-line learning  stage.
According to the original descriptors and  the IME representations of database images,  we estimate the weights of IME layer by ridge regression.
In on-line retrieval stage, we employ the IME layer to map the original representation of query image with ignorable time cost (2 milliseconds per image).

We experiment on five public standard datasets for image retrieval.
The proposed IME layer significantly outperforms related dimension reduction methods and manifold learning methods.
Without post-processing, Our IME layer achieves a  boost in performance of state-of-the-art image retrieval methods with post-processing on most datasets, and needs less computational cost. {\color{blue} Code is available at }\url{https://github.com/XJhaoren/IME_layer}.

\end{abstract}

}

\maketitle
\IEEEdisplaynontitleabstractindextext
\IEEEpeerreviewmaketitle

\section{Introduction}


\IEEEPARstart{O}{ver} the past decades, image retrieval has received  widespread  attention. The representations are shown to be effective for image retrieval~\cite{bow,multiple_bow_ijcv,multiple_bow_pami,soft_bow_cvpr,soft_bow_pami,2012TMM_BOW,2015TMM_FAST,lclc,vlad,2014TMM_VLAD,fv_cvpr,fv_eccv,tri_embed,faemb,rvd}, which are derived by aggregating Scale-Invariant Feature Transform (SIFT)~\cite{sift} features.
After that, image retrieval methods based on Convolutional Neural Network (CNN)~\cite{feature_map} achieve excellent performance~\cite{off_the_shelf,msop,nc,mr,spoc,rmac,crow,2017TMM_VLAD,netvlad,fine_tune_1,fine_tune_2,fine_tune_3,pwa}.
These methods represent a image as the description vector, and sort the Euclidean distances between the feature vectors of query and database images as the retrieval results.

\begin{figure}
  \centering
  \includegraphics[width=3in]{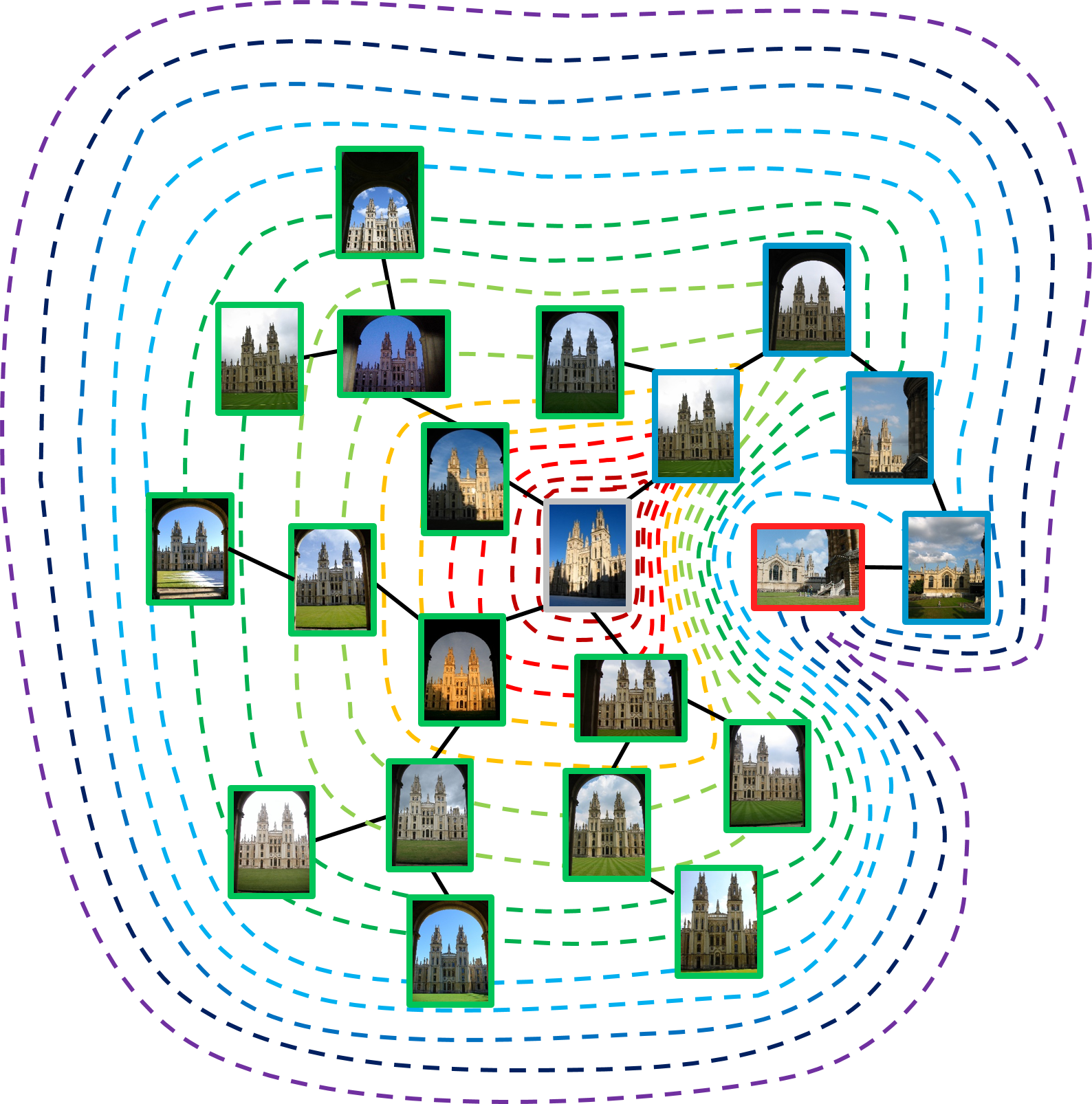}\\
  \caption{Image manifold.
  The image with gray border is the query.
  The Euclidean  distances to query image are expressed by the distance between the positions of the query and reference images in Fig. 1.
  Contour lines correspond to the geodesic distance.
  The images with green and red border are positive samples and negative samples respectively.
  The wires show the neighbourhood relationships of images.
  As shown above, a negative sample  is closer to query than some positive samples in Euclidean space.
  However, it is  far from query image along the wires in manifold.
  The shortest path between the negative sample  and query are shown by images with blue border.}\label{manifold_space}
\end{figure}

Manifolds are the fundamental to perception~\cite{manifold_perception}.
For example, to recognize the faces, the brain equates all images from the same manifold but distinguishes between images from different manifolds.
In image collection, objects and landmarks are depicted in various conditions, such as different viewing angles and under various illumination.
As a consequence, query and relevant images are often connected by a sequence of images, where consecutive images are similar.
The descriptors of these images form a manifold in the descriptor space.
As shown in Fig.~\ref{manifold_space}, some negative samples are far from query image  through the paths in K-NN graph (geodesic distances) but are closer to query than some positive samples based on Euclidean distances.

But the data in image collection is incomplete in most situation.
These images do not vary smoothly.
The data sampled from  intrinsic  manifolds are too sparse.
As a result,  the manifolds reconstructed by these sparse data have some holes and are not continuous and smooth.
The holes lead to the large calculation  error of geodesic distance.
Moreover, existing  manifold learning methods~\cite{isomap,lle,le,sne,t-sne,line,largevis} are not appropriate for image retrieval task, because most  of them are unable to process query image and they have much additional computational cost especially for large scale database.
Except for IsoMap~\cite{isomap} and LLE~\cite{lle}, these manifold learning methods can not handle new image at query time.
And the computational cost  for query image is high for IsoMap and LLE  due to the computation of k-NN of query image.

For the above problems, we propose the iterative manifold embedding (IME) layer to explore the intrinsic manifolds  by incomplete data in this paper.
The weights of the IME layer are learned off-line by unsupervised strategy.
In the on-line query stage,  our IME layer  maps the features of query images into embedding space with very little or even ignorable additional computational complexity.

The IME layer solves the problem of sample holes from two aspects.
(1) The points that share similar neighbours tend to be similar to each other~\cite{line}.
We employ this natural intuition to improve the topological instability of k-NN graph.
The information of second-order proximity suppresses the interference of the sample holes.
(2) We utilize the Euclidean distances to correct the calculation  error of geodesic distances.
Based on the corrected geodesic distances, we embed the data into low-dimensional space and  preserve the intrinsic  geometry of the data.
The above steps are repeated many times to  construct the stable and rubout k-NN graph by incomplete data.
To adapt our algorithm to image retrieval task, we simplify and approximate the IME by linear mapping, called IME layer in this paper.
The IME layer is the integration and simplification version of IME, which reduces the computational cost and estimation error of geodesic distances for query images.
The query image is embedded with very little or even ignorable additional computational cost by IME layer in the on-line retrieval stage.
Working as the additional fully connected layer, the proposed IME layer can be directly connected to  CNNs~\cite{vgg,netvlad,fine_tune_1,fine_tune_2,fine_tune_3}.
For SIFT-based representations~\cite{vlad,fv_cvpr,fv_eccv,tri_embed,faemb,rvd},  IME layer can work as the transform matrix to map the vector representation into low-dimensional space and preserve the original neighbourhood relationships.

We conduct extensive experiments on five public standard  image retrieval datasets, including landmarks and logos.
Experiments results show that our proposed algorithm for manifold-based embedding  significantly improves the performance of global representation vectors.
The proposed IME layer achieves a significant boost in the performance of the related dimension reduction methods and manifold learning methods.
Without reranking, our IME layer still  outperforms the state-of-the-art methods based on search reranking in post-processing step on most datasets.
On a set of five thousand images with 2048 dimensions, IME layer is up to twenty-seven times faster than other manifold learning methods~\cite{isomap,lle}  at query time.
The computational time of IsoMap~\cite{isomap} and LLE~\cite{lle} increases as the scale of database grows, while the cost of our IME layer does not change.
On the large scale dataset that contains 27000  images, our IME layer is more than 120 times faster than other manifold learning methods.
Therefore our IME layer is  efficient and effective  for large scale image retrieval.

The main contributions of this paper are summarized as follows:
\begin{itemize}
  \item We propose a iterative manifold embedding (IME) approach, which explores the intrinsic manifold by incomplete data.
      To suppress the interference of the sample holes, we employ the second-order proximity and original Euclidean distances to correct the geodesic  distances during the iteration process.
      By reconstructing the manifold of database images, our IME method reduces the dimensions of the original representation vectors and  enhances the discrimination of the embedded representations.
  \item We propose the iterative manifold embedding (IME) layer to simplify and accelerate calculation of IME,  which is the integration and simplification version of IME.
      The weights of IME layer are learned off-line  according to original representations and embedded representations by ridge regression.
      With embedding time below 2 milliseconds, the trained iterative manifold embedding layer can be directly connected to CNNs~\cite{vgg,netvlad,fine_tune_1,fine_tune_2,fine_tune_3} or independently work as the transform matrix to map the SIFT-based representations~\cite{vlad,fv_cvpr,fv_eccv,tri_embed,faemb,rvd}.
\end{itemize}

The paper is organized as follows.
In Section~\ref{Related work} we discuss the previous work related to manifold learning and manifold-based methods for image retrieval.
Then, we illustrate the formulation of the proposed algorithm and derive the solution in detail in Section~\ref{The proposed approach}.
The experimental results are described in Section~\ref{Experiment}.
Finally, Section~\ref{Conclusion} concludes the paper and Section~\ref{Future_work} introduces our future work.

\section{Related work}
\label{Related work}
In this section, we review several previous related works from two aspects:  manifold learning methods and manifold-based image retrieval methods.

To the best of our knowledge, very few manifold learning methods can be directly applied to image retrieval.
Instead, most manifold learning methods pay attention to dimension reduction and data visualizations.
Some manifold-based  methods are applied to image retrieval in the search reranking process.
Our IME layer embeds the original representations in image representation process, based on the image manifold reconstructed by incomplete data.

\subsection{Manifold learning}
Our work is related to the manifold learning and dimension reduction methods, such as IsoMap~\cite{isomap}, LLE~\cite{lle}, Laplacian Eigenmap~\cite{le}, SNE~\cite{sne}, t-SNE~\cite{t-sne}, LINE~\cite{line} and LargeVis~\cite{largevis}.

The most popular method for manifold learning may be the IsoMap~\cite{isomap}, which  preserves shortest graph path distance by MDS~\cite{mds} method.
IsoMap first constructs the k-NN graph of data, and then it computes the shortest path distances between all pairs of points according to the k-NN graph.
Finally, the distance vectors are embedded into a low dimensional space.
By exploiting the local symmetries of linear reconstructions, LLE~\cite{lle} is able to learn the global structure of nonlinear manifolds.
Laplacian Eigenmap~\cite{le} constructs a representation for data sampled from a low dimensional manifold embedded in a higher dimensional space  by geometrically motivated algorithm.
SNE~\cite{sne} minimizes the Kullback-Leibler divergences between the original and induced distributions to preserves neighbour identities as well as possible.
After that, the Student t-distribution  is used to solve the crowding problem in t-SNE~\cite{t-sne} instead of the Gaussian distribution in SNE~\cite{sne}.
By exploiting the first-order proximity and the second-order proximity between the vertices, LINE~\cite{line} designs the objective function that preserves both the local and global network structures.
Instead of building a large number of trees to obtain a highly accurate k-NN graph, LargeVis~\cite{largevis} uses neighbour exploring techniques to improve the accuracy of the graph.

These manifold learning methods can not be used directly for image retrieval except for IsoMap~\cite{isomap} and LLE~\cite{lle}, because we can not get the embedded representation for query image.
To map a new query image, IsoMap~\cite{isomap} estimates the geodesic distances between query image and database images by constructed k-NN graph, and then reduces the dimensions of geodesic distances vector.
LLE~\cite{lle} computes the k-NN of a query image and presents the query image by the weighted sum.
The computational cost of the embedded representation for query image is high for IsoMap~\cite{isomap} and LLE~\cite{lle}  due to the computation of k-NN of query image.
The sample holes seriously interfere with the stability of k-NN graph.
Therefore IsoMap~\cite{isomap} and LLE~\cite{lle} are not robust, especially for incomplete data.
Our IME layer embeds the original representation of query image quickly, and is not sensitive to the interference of  sample holes.

\subsection{Manifold-based image retrieval}
Some manifold-based  methods are applied to image retrieval in the context of convolutional features as the post-processing, e.g., query expansion~\cite{qe,2014TMM_qe} and diffusion on region manifold~\cite{diffusion,2016TMM_diffusion,region_manifold}.
Manifold-based methods that leverage  the information of image manifold in the search reranking process are introduced into image retrieval and achieve outstanding performance.

A number of the highly ranked results that satisfy strong spatial constraints from the original query are reissued as a new query in query expansion~\cite{qe} in search reranking process.
The average query expansion (AQE)~\cite{qe} is now used as a standard post-processing of the image retrieval methods, due to its efficiency and significant performance boost.
However, AQE only explores the first-order neighbourhood of query images.
Recursive average query expansion~\cite{qe} further improve the results by explicitly crawling the image manifold, but it increases much cost of query time.
Different with query expansion exploits the manifold of images at query time, diffusion~\cite{diffusion,region_manifold} constructs the neighborhood graph of the dataset off-line and uses this information at query time to search on the manifold.
In the recent work~\cite{region_manifold}, the diffusion on image manifold is used to compute the rerank scoring in the search reranking process.

Working as the post-processing, these manifold-based image retrieval methods need much additional computational cost of reranking at query time.
Without post-processing, our IME layer still achieves better performance than these state-of-the-art image retrieval methods on most datasets.
And the additional computational cost of our IME layer is ignorable (less than 2 milliseconds) in on-line retrieval stage.

\begin{figure*}
  \centering
  \includegraphics[width=7 in]{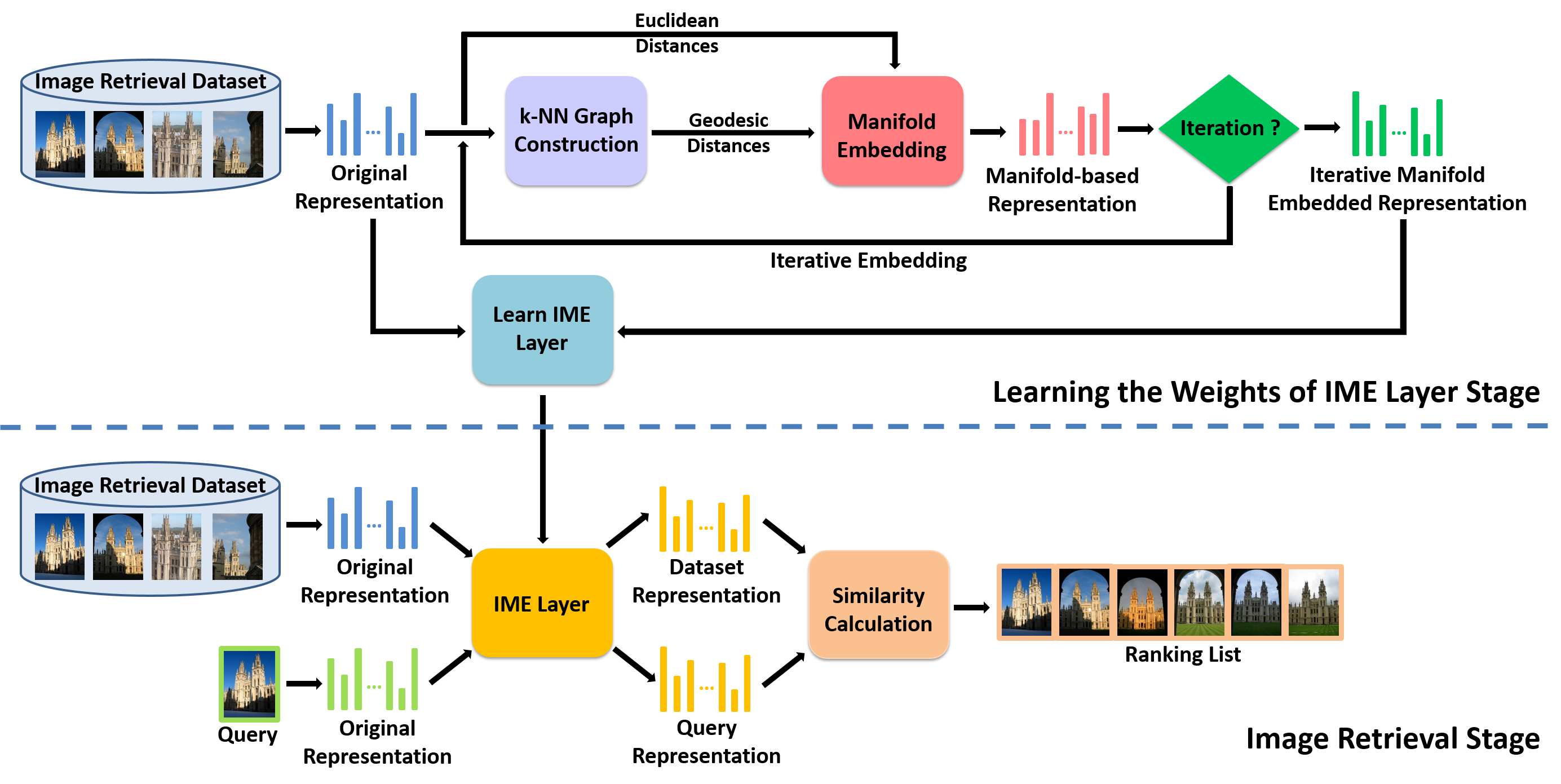}\\
  \caption{Flow chart of the  image retrieval framework which is based on proposed IME layer.
  We learn the weights of iterative manifold embedding (IME) layer in the first off-line stage.
  Then, we use the IME layer to embed the original representations according to the intrinsic manifold rapidly in the on-line image retrieval stage.
  While constructing the k-NN graph, we exploit the information of second-order proximity to suppress  the  interference  of sample holes.
  The Euclidean distances of original representations are utilized to correct  the calculation error of geodesic distances in the manifold embedding step.
  In order to better reconstruct the intrinsic manifold by incomplete sampled data, we repeat the k-NN graph construction and manifold embedding steps many times.
  By adequate approximation and simplification, IME layer embeds the original representations into IME representations and preserves the intrinsic manifold of database images.
}\label{IME}
\end{figure*}

\section{The proposed approach}
\label{The proposed approach}
The diagram of the proposed method is shown in Fig.~\ref{IME}.
We learn the weights of iterative manifold embedding (IME) layer in the first off-line stage.
Then, we embed the original representations into IME representations rapidly by the proposed IME layer in the on-line image retrieval stage.
While constructing the k-NN graph, we exploit the information of second-order proximity to suppress  the  interference  of sample holes.
The Euclidean distances of original representations are utilized to correct  the calculation error of geodesic distances in the manifold embedding step.
In order to better reconstruct the intrinsic manifold by incomplete sampled data, we repeat the k-NN graph construction and manifold embedding steps many times.
By adequate approximation and simplification, IME layer embeds the original representations into IME representations which preserve the intrinsic manifold of database images with ignorable additional time.
IME layer is the integration version of IME, which reduces the estimation error of geodesic distances and the loss of discriminative information in dimension reduction by ridge regression.

In this section, we describe our novel IME layer for image retrieval in detail.
Firstly, we show  the formulation of proposed iterative  embedding method which embeds the original representation according to the intrinsic  manifold of images in Section~\ref{Iterative manifold embedding}.
Then in Section~\ref{IME layer} the proposed embedding  is equivalently implemented by the  fully connected layer, which is called IME layer.

\subsection{Iterative manifold embedding}
\label{Iterative manifold embedding}
The iterative manifold embedding (IME) algorithm has two cyclic steps, which are detailed in  Algorithm.~\ref{alg1}.
In the first step, we construct the k-NN graph and calculate the geodesic distances, considering the information of second-order proximity.
Then, the original representations   are embedded into the manifold-based representations  which preserve the corrected geodesic distances  in the second step.

The first step constructs the k-NN graph $G$ and calculates the  geodesic distances $D_{G}$ of original representations $\psi_{X}=[\psi_{X}(1), \psi_{X}(2), ...,  \psi_{X}(d)]$ (where $\psi_{X}(i)$ is the  n-dimensional original representation vector of image $i$ and $d$ is the database scale).

{\color{blue}
To construct the first-order k-NN graph $G_{1th}$ , each point is only connected to its k nearest neighbours based on the Euclidean distances $D_{X}=\{d_{X}(i,j)\}$ between pairs of images $i, j$ in the input space $X$.
The neighbourhood relations are coarsely represented  as a weighted first-order graph $G_{1th}$ over the data, with the edges of weight $d_{X}(i,j)$ between neighbouring images. The weights $W_{G_{1th}}$ of the edges in first-order graph $G_{1th}$ are defined as the following $d \times d$ matrix:
\begin{equation}\label{7}
{{\mathop{\rm W}\nolimits} _{{G_{1th}}}}(i,j) = \left\{ {\begin{array}{*{20}{c}}
{{d_X}(i,j),\;i~is~connected~to~j}\\
{0,\;i~is~not~connected~to~j}
\end{array}} \right.
\end{equation}
Where $W_{G_{1th}}(i,j)=0$ indicates that the edge between images $i$ and $j$ is cut in first-order graph $W_{G_{1th}}$.
$W_{G_{1th}}(i,j)={d_X}(i,j)$ indicates that image $i$ is connected to $j$ by the edge with path distance ${d_X}(i,j)$.
}

To suppress the interference of the sample holes, we exploit the second-order proximity in the constructed graph $G_{1th}$.
Like the idea in LINE~\cite{line}, we assumes that the vertices sharing many connections to other vertices are similar to each other.
Different with LINE~\cite{line}, we use this natural intuition to construct the stable graph $G$ in a simple but effective way.
The weights $W_{G}$ of the edges in graph $G$ are calculated by the weights $W_{G_{1th}}$ in first-order graph $G_{1th}$:
\begin{equation}\label{0}
{W_G} = {W_{{G_{1th}}}}{W_{{G_{1th}}}}
\end{equation}
Similar to $W_{G_{1th}}(i,j)$, $W_{G}(i,j)=0$ indicates that image $i$ is not connected to $j$ directly in second-order graph $W_{G}$.
The weight $W_{G}(i,j)\neq0$ indicates the distance between directly connected images $i$ and $j$ in second-order graph, which is more robust for sparse samples.
{\color{blue}
In the condition that the samples are sparse, the k nearest neighbours based on the Euclidean distances $D_{X}$ are farfetched and unstable sometimes.
The second-order graph utilizes the natural intuition that the vertices sharing connections to other vertices are similar to each other to ensure the stability of connection.
}

We estimate the geodesic distances $d_{G}(i,j)$  between all pairs of images on the manifold by computing their shortest path distances $d_{G}(i,j)$ in the graph $G$.
The shortest path is simply computed  by Floyd-Warshall algorithm~\cite{floyd}.

Although the information of second-order proximity is utilized to suppress the interference of sample holes, the geodesic distances $d_{G}(i,j)$  still have estimation error.
To further  reduce the error of reconstruction of image manifold, we correct the matrix of geodesic distances $D_{G}=\{d_{G}(i,j)\}$ by Euclidean distances $d_{X}(i,j)$ in the second step.
We embed the representations into a low-dimensional space that best preserves the manifold's estimated intrinsic geometry.
The manifold-based representations  $\psi_{Y}=[\psi_{Y}(1), \psi_{Y}(2), ..., \psi_{Y}(d)]$ (where $\psi_{Y}(i)$ is the manifold-based representation vector of image $i$) are computed by minimizing the cost function
\begin{equation}\label{1}
\begin{split}
& \min \left\| {S({D_G})} \right. + \omega {\left. {S({D_X}) - {\psi _Y}^T{\psi _Y}} \right\|_F}^2
\\
& s.t. ~{\psi _Y}{\psi _Y}^T = I
\end{split}
\end{equation}
where  $\|A\| _{F}$ denotes the $F$-norm of matrix $A$ and the parameter $\omega$ denotes the strength of the correction.
$S(D)$ is a conversion function that converts distances $D$ to similarity.
Given distance matrix $D=\{d(i,j)\}$, many conversion functions can be used such as quadratic function  $S(d(i,j))=- \frac{d(i,j) ^2}{2}$ and t-distribution ${S(d(i,j)) = \frac{1}{{1 + {d(i,j) ^2}}}}$ in practice.
The solution of Eq.~\ref{1} using quadratic function as  conversion function $S$ is equivalent to  the global optimal  solution of the cost function in IsoMap~\cite{isomap}.
We compare different conversion functions in Section~\ref{Experiment}.
In order to retain more discriminatory information, we constrain the embedded representations $\psi_{Y}$ to be orthonormal.

\renewcommand{\algorithmicrequire}{\textbf{Input:}}
\renewcommand{\algorithmicensure}{\textbf{Output:}}
\begin{algorithm*} 
\caption{Iterative Manifold Embedding} 
\label{alg1} 
\begin{algorithmic}[1] 
\REQUIRE Original representation $\psi_{X}$, number of iterations $Iter$.
\ENSURE Iterative manifold embedding representation  $\psi_{IME}$
\STATE $I=0$;
\WHILE{$I<Iter$}
\STATE 1. Construct graph $G$ and calculate the geodesic distances $D_{G}$:
\STATE \qquad Compute Euclidean distances $D_{X}$ of updated representations $\psi_{X}$;
\STATE \qquad Construct first-order neighbour graph $G_{1th}$ based on $D_{X}$ by Eq.~\ref{7};
\STATE \qquad Calculate the weights $W_{G}$ of the edges in graph $G$: ${W_G} = {W_{{G_{1th}}}}{W_{{G_{1th}}}}$;
\STATE \qquad Compute geodesic distances $D_{G}$ by Floyd-Warshall algorithm~\cite{floyd};
\STATE 2. Map the representations $\psi_{X}$ into embedding space that best preserves the manifold’s estimated intrinsic geometry:
\STATE \qquad Correct the geodesic distances $D_{G}$ by Euclidean distances $D_{X}$;
\STATE \qquad Compute d-dimensional representation $\psi_{Y}$ by solving  Eq.~\ref{1};
\STATE $\psi_{X}=\psi_{Y}$;
\STATE  $I++$;
\ENDWHILE
\STATE $\psi_{IME} = \psi_{Y}$;
\end{algorithmic}
\end{algorithm*}

The global minimum of Eq.~\ref{1} is achieved by setting the representations $\psi_{Y}$ to the top $d$ eigenvectors of the similarity matrix
\begin{equation}\label{3}
S_{COR}=S(D_{G})+\omega S(D_{X})
\end{equation}
Let $\lambda_{p}$  be the p-th largest eigenvalue of the matrix $S_{COR}$, and  $v^{i}_{p}$ be the $i$-th component of the $p$-th eigenvector.
Then, the $p$-th component of the  m-dimensional manifold-based representation $\psi_{Y}(i)$ equals to $\sqrt{\lambda_{p}}v^{i}_{p}$.
The dimension $m$  impacts the discrimination of the representation $\psi_{Y}(i)$.
The selection of $m$ is a compromise between  computational cost and accuracy.

The manifold-based  representation $\psi_{Y}(i)$ exploits the neighborhood relationships to represent the feature of a image.
We only use the small amount of neighbours with high credibility to construct k-NN graph in this paper.
The constructed k-NN graph in first step is sparse and has few negative neighbour pairs that represent different objects.
But some important connection paths are cut off.
The sample holes do harm to the reliability of graph $G$.
The correction operations in Eq.~\ref{0} and Eq.~\ref{3}  solve  this problem by exploring the information of second-order proximity and Euclidean distances.
The performance of the embedded representations is improved significantly in this way.

Although these  strategies is straightforward, the remarkable performance is achieved  with some iterations.
We repeat above two steps  to improve the stability of the graph $G$.
{\color{blue}
In each iteration, the Euclidean distances $D_{X}$ are updated based on manifold-based representations $\psi_{Y}$.
In Algorithm.~\ref{alg1}, the iteration process is stated in detail.
}
After a few iterations, we get the final m-dimensional iterative  manifold embedding representations $\psi_{IME}=[\psi_{IME}(1), \psi_{IME}(2), ..., \psi_{IME}(d)]$ (where $\psi_{IME}(i)$ is the IME representation vector of image $i$).
We map the original representations into a low-dimensional space that preserves the geometry of the image manifold.

\subsection{Integration as IME layer}
\label{IME layer}
The  iterative manifold embedding (IME) method proposed above embeds the original representations $\psi_{X}$ into manifold-based representations $\psi_{IME}$.
But if we directly apply IME  for  query image, the embedding leads to more feature extraction time of query images and more estimation error of geodesic distances.
Similar to Isomap~\cite{isomap}, the IME  proposed in previous Section~\ref{Iterative manifold embedding} needs to compute the shortest pathes from a query image to all database images and estimate the geodesic distances  at query time.
{\color{blue}
They are implemented in image retrieval by linking the query into the graph of geodesic distances of the training data.
First the k nearest neighbors of query are found in the training data.
Then, the shortest geodesic distances from query to each point in the training data are computed and transformed into similarity vector by conversion function.
The similarity vector corrected by Euclidean distance is projected into the IME representation by the eigenvector matrix of training data finally.
}
The additional computational cost is proportional to the database scale.
In order to reduce the estimation error and computational cost, the IME method is equivalently implemented and simplified by the fully connected layer in this section, which is called IME layer in this paper.

The IME layer can be regarded as transform matrix which integrates the estimation of geodesic distances with dimension reduction.
We learn the transform matrix according to the original representations $\psi _X$ and the IME representations $\psi _{IME}$ of database images.
We minimize the  following objective function $f(M)$ to  calculate the weights of IME layer as the transform matrix $M\in \Re^{n\times m}$.
\begin{equation}\label{4}
\mathop {{\rm{arg}}min}\limits_M \left\| {{\psi _X}^TM - } \right.\left. {{\psi _{IME}}^T} \right\|_F^2 + \alpha \left\| M \right\|_F^2
\end{equation}
Where $\psi_{X}\in \Re^{n\times d} $ and $\psi_{IME} \in \Re^{m\times d}$ are  original representations and IME representations of the d images on retrieval database respectively. $n$ and $m$ are the dimensions of original representation and IME representation respectively.
$\|M\| _{F}$ denotes the $F$-norm of matrix $M$.
The first term in the objective function is the  transformation  cost term for minimizing the difference between the representations computed by the IME algorithm and  representations mapped by the IME layer.
$\| M \|_{F}$ is a regularization term, and $\alpha$ controls the weight of the regularization.

In this case, this is equivalent to a ridge regression problem and $M$ has a closed form solution.
Let gradient $\frac{\partial f(M)}{\partial M}$=0 to minimize the objective function $f(M)$.
\begin{equation}\label{6}
\frac{\partial f(M)}{\partial M}=({\psi _X}{\psi _X}^T + \alpha I)M - {\psi _X}{\psi _{IME}}^T = 0
\end{equation}
Then, this reduces to
\begin{equation}\label{5}
M=(\psi_{X} \psi_{X}^{T}+\alpha I)^{-1} \psi_{X} \psi_{IME}^{T}
\end{equation}
Where $I$ is the identity matrix.
Since $n$ is the dimensions of original representation, which is low, solving this problem (which needs to be solved only once, at learning time) is extremely fast.

IME layer is the integration and simplification version of IME.
The computational complexity  of IME layer is low at both learning and retrieval steps.
IME layer simplifies the  calculation processes and integrates the estimation of geodesic distances with dimension reduction by ridge regression.
Through the integration of IME, we reduce the estimation error and the loss of discriminative information.
{\color{blue}
In IME, the representation of query image is directly computed by the corrected geodesics distances projected by the eigenvector matrix of training data.
Therefore the estimation error of geodesics distances affect the result significantly.
The performance of IME also depend on the parameters of computing processes very much.
The IME layer is the integration of IME, which diminishes the number of parameters and omits the computation of the shortest geodesic distances from query to each point in the training data.
The cumulative computation error in intermediate step is avoided by integration.
}
In practice, our integration strategy also can be applied to other manifold learning methods~\cite{isomap,lle,le,sne,t-sne,line,largevis} for image retrieval task.
For SIFT-based representations~\cite{vlad,fv_cvpr,fv_eccv,tri_embed,faemb,rvd},  IME layer can work as the transform matrix to map the vector representations into embedding space.
It also can be directly connected to CNNs~\cite{vgg,netvlad,fine_tune_1,fine_tune_2,fine_tune_3} as an additional fully connected layer for CNN-based representations.

\section{Experiment}
\label{Experiment}
This section presents the experimental setup and investigates the accuracy of our approaches for image retrieval on five public datasets.
To evaluate the efficiency and effectiveness of our IME layer, we compare the IME layer with the related manifold learning methods and the state-of-the-art image retrieval methods.
\subsection{Datasets}
We evaluate the performance of our IME layer on  five standard datasets for image retrieval.
Mean average precision (mAP) is used  as the performance measure on all datasets.

Two are well-known image retrieval benchmarks: Oxford5k~\cite{oxford} and Paris6k~\cite{paris}.
Oxford5k contains 5062 images collected from Flickr by searching for particular Oxford landmarks.
Paris6k dataset contains 6412 photographs from Flickr associated with Paris landmarks.
55 queries corresponding to 11 buildings are manually annotated.
The performance is measured using mean average precision (mAP) over the 55 queries.

For large scale image retrieval, we experiment at Oxford105k and Paris106k datasets which add 100k distractor images from Flickr~\cite{oxford}.

The fifth dataset is the recently introduced instance search dataset called INSTRE~\cite{instre}.
It contains various everyday 3D or planar objects from buildings to logos with many variations such as different scales, rotations and occlusions.
Some objects cover a small part of the image, making it a challenging dataset.
It contains of 28543 images from 250 different object classes.
In particular, 100 classes with images retrieved from on-line sources, 100 classes with images taken by the dataset creators, and 50 classes consisting of pairs from the second category.
Different from the original protocol~\cite{instre} that uses all databases images as queries, we evaluate the performance in the same way as the recent works~\cite{region_manifold}.
The INSTRE dataset is randomly split into 1250 queries, 5 per class, and 27293 database images, while a bounding box defines the query region.
The query and the database sets have no overlap.

\subsection{Implementation details}
Our IME layer can combine with both SIFT-based representations and CNN-based representations.
For CNN-based representations, we employ  the  fine-tuned network for image retrieval~\cite{fine_tune_3} to extract the representation vectors.
This fine-tuned ResNet101 produces 2048 dimensional representations.
We extract regions at 3 different scales as in R-MAC~\cite{rmac}, and we additionally include the full image as a region.
In this fashion, each image has  21 regions on average.
The regional representations are aggregated and re-normalized to unit norm in order to construct the original representations, which is exactly as in R-MAC~\cite{rmac}.
For SIFT-based representations, we employ the triangulation embedding~\cite{tri_embed} to aggregate the RootSIFT descriptors~\cite{rootsift}.
In practice, we employ the 8064-dimensional representations of which the vocabulary size is 64.

The  weights of the correction term and regularization term are set as $\omega=2.0$ and $\alpha=1.0$ respectively, throughout our experiments.
Time measurements are reported with a 32-core Intel Xean 2.2GHz CPU.

\subsection{Impact of different components}
In this section, we conduct a series of experiments on second-order proximity, similarity computation,  parameters of IME , IME layer and various original representations.

\textbf{Second-order proximity.} To demonstrate the effectiveness of second-order proximity information, we compare the results of employing first-order graph $G_{1th}$  and graph $G$ respectively to compute the geodesic distances in Table.~\ref{second-order}.
The weights of the edges in graph G are calculated based on both first-order and second-order relationships of database images.
Obviously, the performance of graph $G$ is consistently better on all datasets.
The results show that graph $G$ is more reliable to calculate geodesic distance.
The second-order neighbour relationship is effective to suppress the interference of manifold's sample holes.
As a result, the constructed k-NN graph $G$ is more stable and robust.

\begin{table}[htbp]
  \centering
  \caption{Performance comparison between first-order graph and second-order graph}
    \begin{tabular}{rrrrrr}
    \hline
    \hline
    \multicolumn{1}{c}{\textbf{Graph}}
         & \multicolumn{1}{c}{\textbf{INSTRE}}  & \multicolumn{1}{c}{\textbf{Oxford5k}} & \multicolumn{1}{c}{\textbf{Oxford105k}} & \multicolumn{1}{c}{\textbf{Paris6k}} & \multicolumn{1}{c}{\textbf{Paris106k}}  \\
    \hline
    \multicolumn{1}{c}{\textbf{$G_{1th}$ }} &   \multicolumn{1}{c}{80.7}    &  \multicolumn{1}{c}{89.5}       &   \multicolumn{1}{c}{85.9}      & \multicolumn{1}{c}{92.5}   & \multicolumn{1}{c}{85.6}  \\
    \hline
    \multicolumn{1}{c}{\textbf{$G$}} &   \multicolumn{1}{c}{\textbf{82.4}}    &    \multicolumn{1}{c}{\textbf{92.0}}     &      \multicolumn{1}{c}{\textbf{87.2}}   &  \multicolumn{1}{c}{\textbf{96.6}}   & \multicolumn{1}{c}{\textbf{93.3}} \\
    \hline
    \hline
    \end{tabular}%
  \label{second-order}%
\end{table}%

\begin{figure*}
  \centering
  \includegraphics[width=5 in]{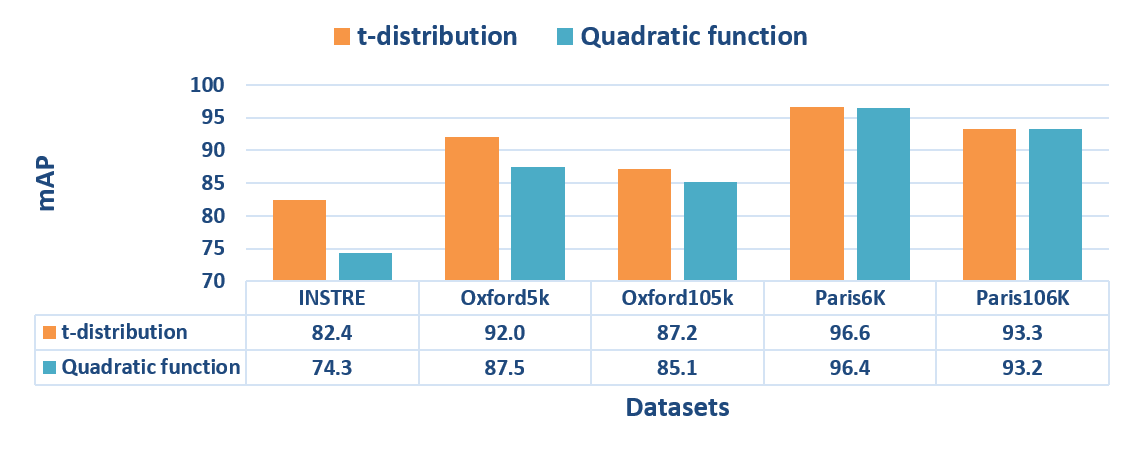}\\
  \caption{Performance comparison of  different methods to convert distances into similarity. T-distribution has better performance, because it reduces the calculation error of similarity between the points that are far apart and enhances the discrimination of moderate geodesic distances.
  }\label{similarity}
\end{figure*}
\textbf{Similarity computation.} We compare different conversion functions in Fig.~\ref{similarity}, such as quadratic function  $S(d(i,j))=- \frac{d(i,j) ^2}{2}$ and t-distribution ${S(d(i,j)) = \frac{1}{{1 + {d(i,j) ^2}}}}$.
These conversion operators are used in IsoMap~\cite{isomap} and t-SNE~\cite{t-sne} respectively.

Fig.~\ref{similarity} shows the performance versus  different functions to convert distances into similarity.
The t-distribution performs much better than quadratic function on all datasets.
It enhances the discrimination of the moderate geodesic distances and suppresses the difference of large geodesic distances.
Due to error accumulation, the calculation error of large geodesic distances is large.
The t-distribution  reduces the calculation error of similarity between  the points that are far apart by suppressing the difference of large geodesic distances, while quadratic function increases the calculation error.
Therefore, we employ a Student t-distribution with one degree of freedom (which is the same as Cauchy distribution)  as the conversion function in our IME layer.

\textbf{Parameters of IME.} 
We evaluate the performance of  different parameters of IME, such as the number of iterations $Iter$, dimensions of IME representation $m$, the number of neighbours $k$ and  the strength of correction $\omega$.

\begin{figure*}
  \centering
  \includegraphics[width=7 in]{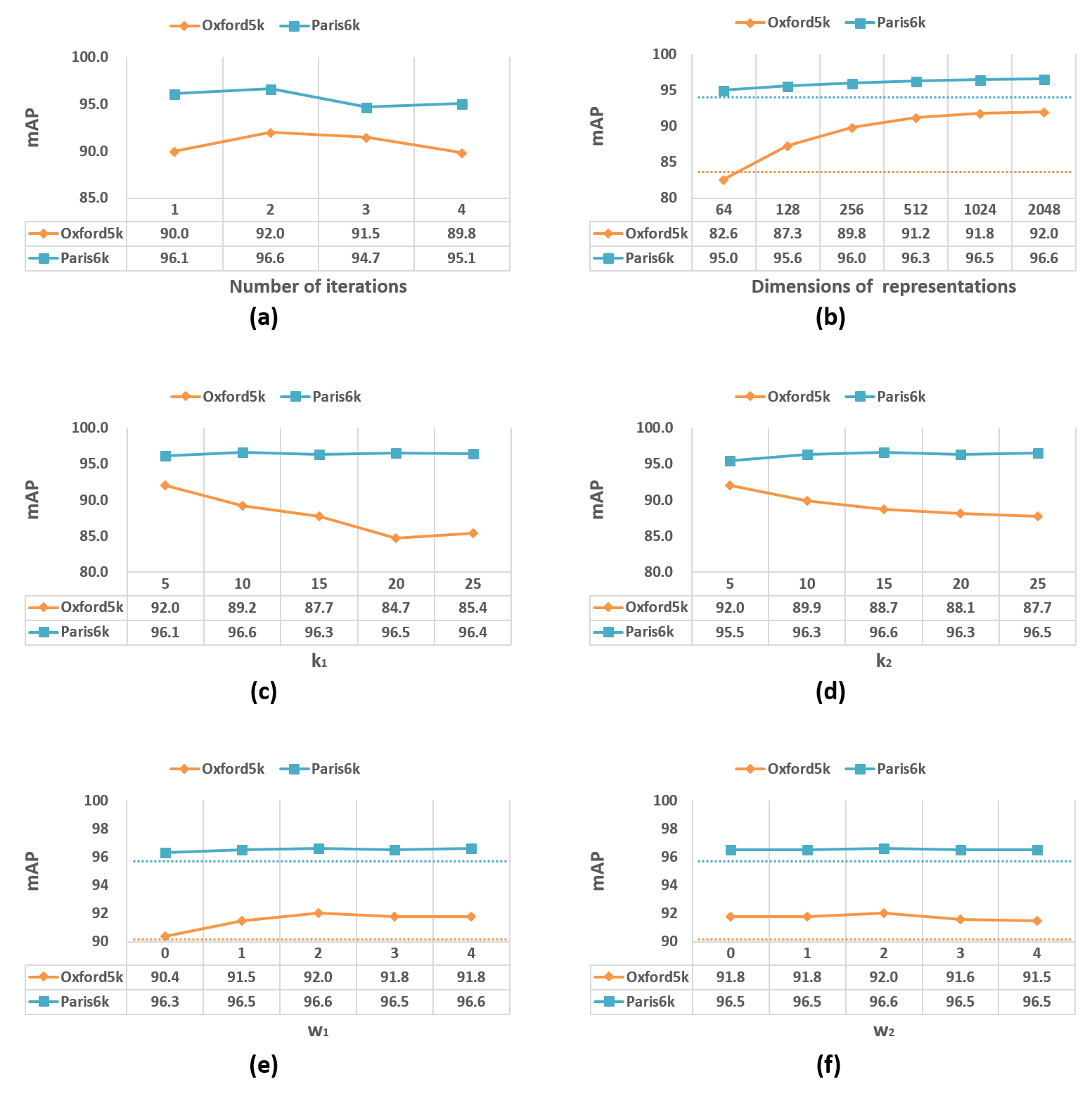}\\
  \caption{Performance comparison of  different parameters on three datasets with varying the number of iterations, dimensions of IME representation,  number of neighbours and  correction weights.
  The number of iterations are set as $Iter$=2 due to its better performance.
  Even if the dimension $m$ is reduced to 128, the performance of IME representation is better than original representation.
  The mAP of original representation is shown in (b) by dotted lines.
  Our method is not heavily relied on the correction weights $w$, and we set $w$=2 in other experiments.
  }\label{IME_parameters}
\end{figure*}

Fig.~\ref{IME_parameters}(a) shows the performance versus the number of iterations $Iter$.
Our IME layer performs well just with a small number of iterations.
{\color{blue}
Too many  iterations lead to the overfitting of embedding.
}
We set $Iter$=2 in the rest experiments in this paper due to the better performance and moderate training time.

The performance comparison of various dimensions of embedded representations is shown in Fig.~\ref{IME_parameters}(b).
We achieve 92.0 and 96.6 on Oxford5k and Paris6k  datasets respectively when we employ the 2048-dimensional IME representation.
{\color{blue}
The high-dimensional representation preserves more discriminative information, so performance of it is better than low-dimensional representation.
}
As shown in Fig.~\ref{IME_parameters}(b) by dotted lines, the mAP of original representation are 83.9 and 93.8  on Oxford5k and Paris6k datasets respectively.
Our method achieves better performance than original representation even if the dimensions of final IME representation is one sixteenth of original representation, 128 dimension.
The results show that our low-dimensional IME representation still preserves the intrinsic manifold.
The reconstructed manifold is effective for image retrieval task to search the similar images.

Fig.~\ref{IME_parameters}(c,d) shows the performance versus the number of neighbours to construct the k-NN graph.
$k_{1}$ and $k_{2}$ are the number of neighbours for first and second iterations respectively.
{\color{blue}
The selection of parameter $k_{i}$ depends on the sparsity of datasets to some degree.
The small $k_{i}$ is better suited for sparser dataset.
With small $k_{i}$, most of the neighbours are positive.
With large $k_{i}$, there are many negative neighbours in the graph especially for the sparse dataset.
Therefore the constructed k-NN graph with too large $k_{i}$ is unstable for sparse dataset.
The calculation error of geodesic distances computed according to the graph with too large $k_{i}$  is large.
But the dense dataset is more adaptable.
Even with large $k_{i}$, the stability of neighbours in dense dataset are little influenced by noise due to more stable neighbours.
There are more similar objects in Paris6k than Oxford6k.
The images in Oxford5k are sparser.
But the images in Paris6k are denser.
Therefore Paris6k is almost unaffected by $k_{i}$ but Oxford5k degrades with $k_{i}$.
}

We evaluate the effect of  various correction weights $\omega$, and then report the results in Fig.~\ref{IME_parameters}(e,f).
$\omega_{1}$ and $\omega_{2}$ are the strength of correction for first and second iterations respectively.
The results show that the performance of our method does not heavily rely on the correction weights $\omega$.
We set the correction weights $\omega=2.0$ in all the other experiments.
{\color{blue}
The estimation error of geodesic distance is larger in the first iteration.
$\omega_{1}$ has a more significant impact than $\omega_{2}$.
}
If we do not correct the geodesic distances $D_{G}$ by Euclidean distances $D_{X}$  while iterative embedding (that is, {\color{blue} $\omega_{i}=0, i=1,2$}), the results are 89.9 and 95.7 on Oxford5k and Paris6k datasets respectively {\color{blue} as shown in Fig.~\ref{IME_parameters}(e,f) by dotted lines}.
datasets respectively.
{\color{blue}
The mAPs of uncorrected geodesic distances ($\omega_{i}=0, i=1,2$) are lower than mAPs of corrected geodesic distances ($\omega_{i}=2, i=1,2$), 92.0 and 96.6 on Oxford5k and Paris6k  datasets respectively.
}
The results demonstrate that the correction is important for the geodesic distances computed by the incomplete data.

\begin{table*}[htbp]
  \centering
  \caption{Performance comparison with other manifold learning methods for image retrieval}
    \begin{tabular}{rrrrrrrrr}
    \hline
    \hline
          & \multicolumn{3}{c}{\textbf{mAP}} &  &\multicolumn{3}{c}{\textbf{Average additional query time (second)}} & \bigstrut \\
          \cline{2-4}
          \cline{6-8}
          \multicolumn{1}{c}{\textbf{Method}}
          & \multicolumn{1}{c}{\textbf{Oxford5k}} & \multicolumn{1}{c}{\textbf{Paris6k}} & \multicolumn{1}{c}{\textbf{INSTRE}} &  &\multicolumn{1}{c}{\textbf{Oxford5k}} & \multicolumn{1}{c}{\textbf{Paris6k}} & \multicolumn{1}{c}{\textbf{INSTRE}} \bigstrut \\
     \hline
     \multicolumn{1}{c}{{PCA~\cite{pca} }}   &  \multicolumn{1}{c}{82.6}    &\multicolumn{1}{c}{91.5}       &\multicolumn{1}{c}{62.2}&       &\multicolumn{1}{c}{0.002}  &\multicolumn{1}{c}{0.002} &\multicolumn{1}{c}{0.002}\\
    \multicolumn{1}{c}{{IsoMap~\cite{isomap} }}   &  \multicolumn{1}{c}{77.9}    &\multicolumn{1}{c}{91.8}       &\multicolumn{1}{c}{68.6}&       &\multicolumn{1}{c}{0.378}  &\multicolumn{1}{c}{0.403} &\multicolumn{1}{c}{3.483}\\
    \multicolumn{1}{c}{{LLE~\cite{lle}}}       &  \multicolumn{1}{c}{51.7}    &\multicolumn{1}{c}{40.5}       &\multicolumn{1}{c}{42.7}&       &\multicolumn{1}{c}{0.054}  &\multicolumn{1}{c}{0.066} &\multicolumn{1}{c}{0.249}\\
    \multicolumn{1}{c}{{IME }}      &  \multicolumn{1}{c}{83.5}    &\multicolumn{1}{c}{93.4}       &\multicolumn{1}{c}{75.9}&       &\multicolumn{1}{c}{0.907}  &\multicolumn{1}{c}{0.937} &\multicolumn{1}{c}{7.659}\\
    \multicolumn{1}{c}{{IME layer}} &  \multicolumn{1}{c}{\textbf{92.0}}    &\multicolumn{1}{c}{\textbf{96.6}}       &\multicolumn{1}{c}{\textbf{82.4}}&       &\multicolumn{1}{c}{\textbf{0.002}}  &\multicolumn{1}{c}{\textbf{0.002}} &\multicolumn{1}{c}{\textbf{0.002}}\\
   \hline
   \hline
    \end{tabular}%

  \label{map_time}%
\end{table*}%

\textbf{IME layer.}
The proposed IME layer is the integration and simplification  version of IME.
{\color{blue}
The accuracy and average additional query time of PCA, IME, IME layer and other manifold learning methods are shown in Table~\ref{map_time}.
IME layer achieves better performance on both mAP and time cost.
The additional computation cost of IME layer is roughly identical to PCA.
On Oxford5k and Paris6k datasets, our IME layer is more than  twenty-seven times faster than IsoMap~\cite{isomap} and LLE~\cite{lle}, the manifold learning methods that can be applied to image retrieval, and significantly outperforms them on mAP.
The computational cost of IME layer is unrelated to the scale of database, while the cost of IsoMap~\cite{isomap} and LLE~\cite{lle} is proportional to the number of database images.
On a large scale dataset INATRE, our method is more than 120 times faster than them.
}
The performance of IME layer is better than IME.
Because the integration  reduces  the calculation error of the geodesics distances between query image and database images and  the loss of discriminative information in dimension reduction.
The results demonstrate that our IME layer is effective and efficient for image retrieval.

\textbf{Various original representations.}
We do experiments on both SIFT descriptors and CNN  features.
Table.~\ref{sift} presents  the results of the accuracy of SIFT-based and CNN-based representations  with/without  IME layer.
The dimensions of final IME representation is reduced to 2048 in this experiment.
The results demonstrate that our IME layer is effective for various features.
The IME layer can be directly connected with a CNN as the trained fully connected layer.
For other features,  our IME layer can work as the transform matrix to map the aggregated representations into embedding space.

\begin{table}[htbp]
  \centering

  \caption{Performance of IME layer for various features}
    \begin{tabular}{rrr}
    \hline
    \hline
     \multicolumn{1}{c}{\textbf{Feature}}     & \multicolumn{1}{c}{\textbf{Oxford5k}} & \multicolumn{1}{c}{\textbf{Oxford105k}}  \\
    \hline
    \multicolumn{1}{c}{{SIFT~\cite{tri_embed}}} &   \multicolumn{1}{c}{52.7}      &    \multicolumn{1}{c}{27.6}         \\
    \multicolumn{1}{c}{{SIFT+IME layer}} &    \multicolumn{1}{c}{\textbf{62.2}}      &    \multicolumn{1}{c}{\textbf{31.3}}       \\
    \hline
    \multicolumn{1}{c}{{CNN~\cite{fine_tune_3}}} &    \multicolumn{1}{c}{83.9}      &    \multicolumn{1}{c}{80.8}       \\
    \multicolumn{1}{c}{{CNN+IME layer}} &    \multicolumn{1}{c}{\textbf{92.0}}      &    \multicolumn{1}{c}{\textbf{87.2}}        \\
    \hline
    \hline
    \end{tabular}
  \label{sift}%
\end{table}%

\begin{table*}[htbp]
  \centering
  \caption{Performance comparison with the state-of-the-art methods.}
    \begin{tabular}{rrrrrrr}
    \hline
    \hline
    \textbf{} & \textbf{} & \multicolumn{5}{c}{\textbf{Datasets}} \bigstrut \\
    \cline{3-7}
    \multicolumn{1}{c}{\textbf{Method}} & \multicolumn{1}{c}{\textbf{Dimensions}} & \multicolumn{1}{c}{\textbf{INSTRE}} & \multicolumn{1}{c}{\textbf{Oxford5k}} & \multicolumn{1}{c}{\textbf{Oxford105k}} & \multicolumn{1}{c}{\textbf{Paris6k}} & \multicolumn{1}{c}{\textbf{Paris106k}} \bigstrut \\
    \hline
    \multicolumn{7}{c}{\textbf{Original representations}} \bigstrut \\
    \hline
    \multicolumn{1}{c}{CroW~\cite{crow}} & \multicolumn{1}{c}{512} & \multicolumn{1}{c}{--} & \multicolumn{1}{c}{68.2} & \multicolumn{1}{c}{63.2} & \multicolumn{1}{c}{79.8} & \multicolumn{1}{c}{71.0}\\
    \multicolumn{1}{c}{R-MAC~\cite{fine_tune_1}} & \multicolumn{1}{c}{512} & \multicolumn{1}{c}{47.7} & \multicolumn{1}{c}{77.7} & \multicolumn{1}{c}{70.1} & \multicolumn{1}{c}{84.1} & \multicolumn{1}{c}{76.8}\\
    \multicolumn{1}{c}{R-MAC~\cite{fine_tune_3}} & \multicolumn{1}{c}{2048} & \multicolumn{1}{c}{62.6} & \multicolumn{1}{c}{83.9} & \multicolumn{1}{c}{80.8} & \multicolumn{1}{c}{93.8} & \multicolumn{1}{c}{89.9}\\
    \hline
    \multicolumn{7}{c}{\textbf{Dimension reduction and manifold learning}} \bigstrut \\
    \hline
    \multicolumn{1}{c}{PCA~\cite{pca}} & \multicolumn{1}{c}{512} & \multicolumn{1}{c}{50.0} & \multicolumn{1}{c}{78.2} & \multicolumn{1}{c}{74.7} & \multicolumn{1}{c}{91.0} & \multicolumn{1}{c}{85.4}\\
    \multicolumn{1}{c}{ICA~\cite{ica}} & \multicolumn{1}{c}{512} & \multicolumn{1}{c}{50.3} & \multicolumn{1}{c}{77.5} & \multicolumn{1}{c}{73.7} & \multicolumn{1}{c}{90.8} & \multicolumn{1}{c}{85.2} \\
    \multicolumn{1}{c}{IsoMap~\cite{isomap}} & \multicolumn{1}{c}{512} & \multicolumn{1}{c}{69.7} & \multicolumn{1}{c}{77.8} & \multicolumn{1}{c}{64.7} & \multicolumn{1}{c}{91.8} & \multicolumn{1}{c}{69.6}\\
    \multicolumn{1}{c}{LLE~\cite{lle}} & \multicolumn{1}{c}{512} & \multicolumn{1}{c}{60.2} & \multicolumn{1}{c}{64.0} & \multicolumn{1}{c}{47.6} & \multicolumn{1}{c}{50.7} & \multicolumn{1}{c}{21.7}\\
    \multicolumn{1}{c}{ IME layer} & \multicolumn{1}{c}{512} & \multicolumn{1}{c}{\textbf{83.1}} & \multicolumn{1}{c}{\textbf{91.2}} & \multicolumn{1}{c}{\textbf{85.1}} & \multicolumn{1}{c}{\textbf{96.3}} & \multicolumn{1}{c}{\textbf{92.5}} \\
\\
    \multicolumn{1}{c}{PCA~\cite{pca}} & \multicolumn{1}{c}{1024} & \multicolumn{1}{c}{58.7} & \multicolumn{1}{c}{80.8} & \multicolumn{1}{c}{78.0} & \multicolumn{1}{c}{91.7} & \multicolumn{1}{c}{86.8}\\
    \multicolumn{1}{c}{ICA~\cite{ica}} & \multicolumn{1}{c}{1024} & \multicolumn{1}{c}{58.7} & \multicolumn{1}{c}{81.6} & \multicolumn{1}{c}{78.0} & \multicolumn{1}{c}{92.1} & \multicolumn{1}{c}{87.3} \\
    \multicolumn{1}{c}{IsoMap~\cite{isomap}} & \multicolumn{1}{c}{1024} & \multicolumn{1}{c}{69.1} & \multicolumn{1}{c}{78.1} & \multicolumn{1}{c}{65.4} & \multicolumn{1}{c}{92.1} & \multicolumn{1}{c}{72.1}\\
    \multicolumn{1}{c}{LLE~\cite{lle}} & \multicolumn{1}{c}{1024} & \multicolumn{1}{c}{50.4} & \multicolumn{1}{c}{58.8} & \multicolumn{1}{c}{42.3} & \multicolumn{1}{c}{45.0} & \multicolumn{1}{c}{16.2}\\
    \multicolumn{1}{c}{ IME layer} & \multicolumn{1}{c}{1024} & \multicolumn{1}{c}{\textbf{82.8}} & \multicolumn{1}{c}{\textbf{91.8}} & \multicolumn{1}{c}{\textbf{86.2}} & \multicolumn{1}{c}{\textbf{96.5}} & \multicolumn{1}{c}{\textbf{92.9}} \\
\\
    \multicolumn{1}{c}{PCA~\cite{pca}} & \multicolumn{1}{c}{2048} & \multicolumn{1}{c}{62.2} & \multicolumn{1}{c}{82.6} & \multicolumn{1}{c}{79.1} & \multicolumn{1}{c}{91.5} & \multicolumn{1}{c}{86.5}\\
    \multicolumn{1}{c}{ICA~\cite{ica}} & \multicolumn{1}{c}{2048} & \multicolumn{1}{c}{62.2} & \multicolumn{1}{c}{82.7} & \multicolumn{1}{c}{79.1} & \multicolumn{1}{c}{91.5} & \multicolumn{1}{c}{86.5} \\
    \multicolumn{1}{c}{IsoMap~\cite{isomap}} & \multicolumn{1}{c}{2048} & \multicolumn{1}{c}{68.6} & \multicolumn{1}{c}{77.9} & \multicolumn{1}{c}{68.3} & \multicolumn{1}{c}{91.8} & \multicolumn{1}{c}{76.4}\\
    \multicolumn{1}{c}{LLE~\cite{lle}} & \multicolumn{1}{c}{2048} & \multicolumn{1}{c}{42.7} & \multicolumn{1}{c}{51.7} & \multicolumn{1}{c}{34.9} & \multicolumn{1}{c}{40.5} & \multicolumn{1}{c}{14.7}\\
    \multicolumn{1}{c}{ IME layer} & \multicolumn{1}{c}{2048} & \multicolumn{1}{c}{\textbf{82.4}} & \multicolumn{1}{c}{\textbf{92.0}} & \multicolumn{1}{c}{\textbf{87.2}} & \multicolumn{1}{c}{\textbf{96.6}} & \multicolumn{1}{c}{\textbf{93.3}} \\
    \hline
    \multicolumn{7}{c}{\textbf{Search reranking}} \bigstrut \\
    \hline
    \multicolumn{1}{c}{QE~\cite{qe}} & \multicolumn{1}{c}{2048} & \multicolumn{1}{c}{70.5} & \multicolumn{1}{c}{89.6} & \multicolumn{1}{c}{\textbf{88.3}} &  \multicolumn{1}{c}{95.3} & \multicolumn{1}{c}{92.7} \\
    \multicolumn{1}{c}{SCSM~\cite{scsm}} & \multicolumn{1}{c}{2048} & \multicolumn{1}{c}{71.4} & \multicolumn{1}{c}{89.1} & \multicolumn{1}{c}{87.3} & \multicolumn{1}{c}{95.4} & \multicolumn{1}{c}{92.5} \\
    \multicolumn{1}{c}{Diffusion~\cite{region_manifold}} & \multicolumn{1}{c}{2048} & \multicolumn{1}{c}{80.5} & \multicolumn{1}{c}{87.1} & \multicolumn{1}{c}{86.8} & \multicolumn{1}{c}{96.5} & \multicolumn{1}{c}{\textbf{95.4}}\\
    \multicolumn{1}{c}{IME layer} & \multicolumn{1}{c}{2048} & \multicolumn{1}{c}{\textbf{82.4}} & \multicolumn{1}{c}{\textbf{92.0}} & \multicolumn{1}{c}{87.2} & \multicolumn{1}{c}{\textbf{96.6}} & \multicolumn{1}{c}{93.3}\\

    \hline
    \hline
    \end{tabular}%
  \label{best}%
\end{table*}%

\begin{figure*}
  \centering
  \includegraphics[width=6 in]{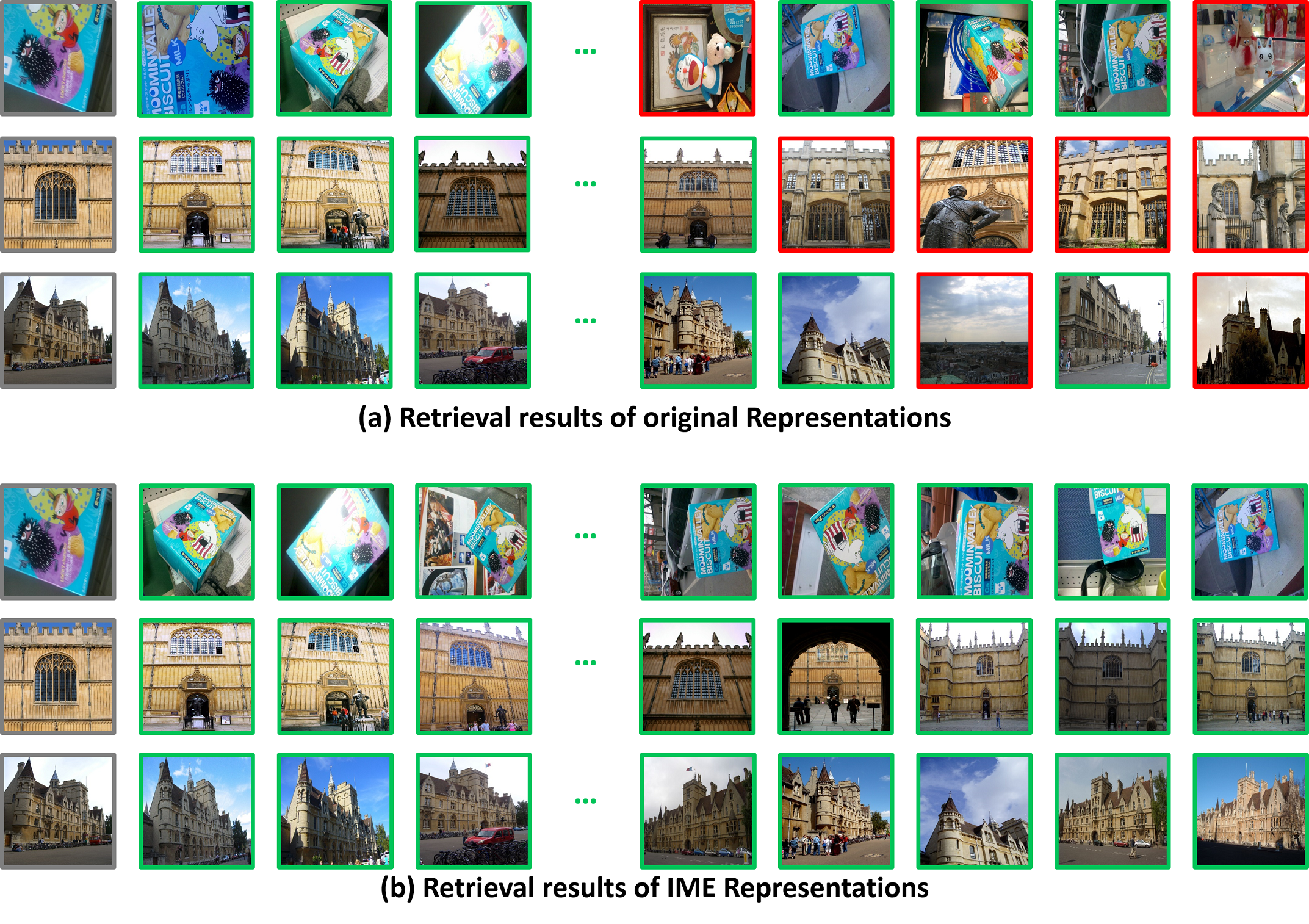}\\
  \caption{Sample retrieval results of original representation and IME representation.
  The images with gray border are query images. The images with green border are positive retrieval results (ground-truth) and the images with red border are negative results.
   Because the reconstructed manifold of images is continuous and smooth, the images that contain the same objects in different viewing angles and various illumination are closer in the embedding space.
  }\label{example}
\end{figure*}
\subsection{Comparison with the state-of-the-art}
We compare with the state-of-the-art approaches with global representation.
Table~\ref{best} summarizes the results.
Our IME layer significantly outperforms all the existing dimension reduction and manifold learning methods on all datasets.
Without post-processing, our IME layer still outperforms the state-of-the-art image retrieval methods with reranking on most datasets.

In the first part of the table,  we show results of the methods that employ global representations of images and do not perform any form of spatial verification or query expansion at query time.
The 2048-dimensional R-MAC vectors~\cite{fine_tune_3} in the first part are employed as the original CNN-based  representations in this paper.

We compare our IME layer with related dimension reduction and manifold learning methods in the second part of the table.
Except for IsoMap~\cite{isomap} and LLE~\cite{lle}, other nonlinear manifold learning methods can not be directly applied to image retrieval.
We consistently outperform  them for various dimensions on all datasets.
In one case (namely, on INSTRE), our method is more than 13 mAP points ahead of the best competitor~\cite{isomap}.
Our IME layer requires less computational cost  compared with IsoMap~\cite{isomap} and LLE~\cite{lle}  in the on-line image retrieval stage, and same computational cost as PCA~\cite{pca} and ICA~\cite{ica}.

As shown in the third  part of the table, we show the results of state-of-the-art methods that employ global representations  and perform  search reranking (e.g. , spatial verification~\cite{scsm}, query expansion (QE)~\cite{qe} or diffusion~\cite{region_manifold}) at query time.
Without reranking, our method still outperforms the state-of-the-art methods with post-processing on most datasets.
These methods with post-processing~\cite{scsm, qe, region_manifold} contain image scoring and ranking steps more than once.
It is worth noting that our IME layer method is much faster than these methods and has comparable performance.

{\color{blue}
For Oxford105k and Paris106k datasets, we use the incomplete data (the images in Oxford5k and Paris6k respectively) and 5000 noisy data from 100k distractor images on Flickr~\cite{oxford} to learn the weights of IME layer in the off-line stage taking a few minutes.
}
In comparison, diffusion~\cite{region_manifold} employs all 100 thousand images to construct the manifold and takes many hours.
Our IME layer has less than 2 milliseconds additional time per query in the on-line retrieval stage, while diffusion~\cite{region_manifold} requires about 14 seconds.
Employing sparse samples to learn the weights of IME layer, our IME layer still achieves good performance on large scale image retrieval.
The results demonstrate that our IME layer is effective to reconstruct image manifold by incomplete data.

In Fig.~\ref{example} we present some query examples using original representation and IME representation respectively.
The images with gray border are cropped query images.
The images with green border are positive retrieval results (ground-truth) and the images with red border are negative results.
IME layer significantly improves the retrieval results by mapping the original representations into embedding space.
Due to the reconstructed continuous manifold of images, the images that contain the same objects in different viewing angles and various illumination are closer in the embedding space.

\section{Conclusion}
\label{Conclusion}
In this paper we propose a manifold learning method called iterative manifold embedding (IME) layer and demonstrate its efficiency and effectiveness for image retrieval.
Through the unsupervised strategy, the weights of IME layer are learned by incomplete data.

Our IME layer introduces the manifold learning into image retrieval.
We solve the sample holes problem on manifold learning, using the information of second-order proximity and the correction  of geodesic distances  by Euclidean distances.
In order to reduce the additional computational cost and estimation error of geodesic distances  at query time, we integrate the manifold-based embedding by the approximate linear mapping.

Experiments on five standard retrieval datasets demonstrate that our IME layer significantly outperforms related dimension reduction and manifold learning methods
with equivalent or lower computational complexity.
Without search reranking, our method still outperforms the state-of-the-art methods with search reranking on most datasets.

{\color{blue}
\section{Future Work}
\label{Future_work}
The main limitation of our IME layer is that the off-line learning is time-consuming. Especially for large datasets, the cost of construction of k-NN graph and calculation of geodesic distances is large. We will try to speed up the off-line learning stage in the future work by parallel computing and other strategies.
The IME layer is learned via a linear transform in this work.
We will employ nonlinear transform and try to learn more layers end-to-end in the future work.
}

\section*{Acknowledgment}
This work was supported by the National Natural Science Foundation of China under Grant 61531019, Grant 61601462, and Grant 71621002.
The authors would like to thank the Associate Editor and the anonymous reviewers for their contributions to improve the quality of this paper.

\ifCLASSOPTIONcaptionsoff
  \newpage
\fi

\bibliographystyle{IEEEtran}
\bibliography{iterative_region_manifold}
%

%
%
%




\end{document}